# Artificial Intelligence Applications in Horizon Scanning for Infectious Diseases

Ian Miles[1], Mayumi Wakimoto[2], Wagner Meira Jr[3]., Daniela Paula[4], Daylene Ticiane[2], Bruno Rosa[2], Jane Biddulph[5], Stelios Georgiou[6], Valdir Ermida[2] †

1. The University of Manchester; 2. Evandro Chagas National Institute of Infectious Diseases/Fiocruz; 3. Federal University of Minas Gerais; 4. National School of Statistical Science; 5. University College London; 6. UK Health Security Agency

## Abstract

This review explores the integration of Artificial Intelligence (AI) into Horizon Scanning (HS) processes, with a focus on identifying and responding to emerging threats and opportunities in the field of Infectious Diseases (IDs). Drawing on the VigIA project, we examine how AI tools—particularly machine learning and generative models—can enhance signal detection, data monitoring, scenario analysis, and decision support. We also address the risks associated with AI adoption and propose strategies for effective implementation and governance. The findings contribute to the growing body of Foresight literature by demonstrating the potential and limitations of AI in Public Health preparedness.

## 1. Introduction

Horizon Scanning (HS) is a Foresight method aimed at identifying early signals of change, including threats and opportunities (Cuhls, 2020; Miles & Saritas, 2012). In the context of Infectious Diseases (IDs), HS can inform Public Health strategies by anticipating outbreaks, evaluating preparedness, and guiding policy responses. The rapid evolution of Artificial Intelligence (AI) presents new possibilities for enhancing HS processes (Bengston et al., 2024; Miles et al., 2016).

---

† Corresponding author: valdir.ermida@fiocruz.br

This narrative review is about the application of AI across the stages of HS, drawing insights from the VigIA Project, coordinated by INI/Fiocruz at Rio de Janeiro, Brazil. INI is the Evandro Chagas National Institute of Infectious Diseases, the clinical research Unit at Oswaldo Cruz Foundation (Fiocruz), the Brazilian Public Research Organisation in the health field.

In connection with IDs, the threats may simply involve the plausible recurrence of known problems, such as new variants of seasonal influenza, or represent very new threats, ranging from quite novel zoonotic viral infections to things as extreme as new disease agents. In addition to the epidemiological threats, there may be threats in terms of preparedness (e.g. shortage of Personal Protective Equipment or hospital facilities) or related to social circumstances (e.g. support for vaccination programmes, distrust of Public Health authorities). Such themes will often feature in risk analysis and risk management activities. Opportunities of various sorts may also be apparent, for instance in the development of improved methods of disease detection and identification, new vaccines and approaches to vaccine development, other preventive measures and treatments for diseases, and policy support for enhancing both research capabilities and public awareness connected with the topic (Brisse et al., 2020; Marmot et al., 2010; Wei et al., 2025).

We may be able to model the course of a disease outbreak when data on characteristics of the disease agent (e.g. its transmissibility across people) have become apparent (Grassly & Fraser, 2008; Kucharski et al., 2020; Willem et al., 2017). But we cannot predict what lies over the horizon: what new diseases may emerge, what new treatments will be available to address these threats. However, it is possible to examine early warning signals that suggest that threats and opportunities are liable to emerge. It may also be possible to specify conditions under which new threats and/or opportunities may become more probable or more significant. Horizon Scanning is one of the tools of Foresight studies that has been developed to address such topics (Amanatidou et al., 2012; Cuhls, 2020; Ravetz et al., 2011; WHO, 2022). The aim of this review is to examine the scope for applying Artificial Intelligence (AI) to HS in this field.

## 2. Systematic Horizon Scanning

Horizon Scanning involves exploring dimensions that surpass immediate contingencies to consider what phenomena may develop in the future that go beyond the immediately predictable (Amanatidou et al., 2012; Cuhls, 2020; Miles et al., 2016). It is undertaken informally by top managers and experts within most organisations – who are interested in their organisation, its operating environment and technologies, and are on the look-out for changes that may disrupt business-as-usual. They follow various news and trade media in order to stay abreast of relevant events, and use their own knowledge to assess whether any of these events points to the possibility, and especially the likelihood, of these events portending significant change.

In periods and fields of activity where there is much ongoing change, it is common for larger organisations to systematise HS, at the very least by inviting key staff to share their perceptions of significant events (these events are treated as signals of possible change). "Weak Signals" are those that are considered to be ambiguous in terms of what and/or how much change they indicate (Miles et al., 2016; Ravetz et al., 2011). More systematic HS is undertaken when organisations institute an HS function, either employing outside agencies or their own staff to undertake an HS exercise, to produce a series of regular HS reports, or to raise issues significant for risk management as they arise (Georghiou et al., 2008; Miles et al., 2016).

Systematic HS typically involves various stages (Georghiou et al., 2008; Miles et al., 2016; Ravetz et al., 2011) :

1. The establishment of an HS framework, or revision of an existing one – which necessarily means some specification of the focal objects or topics for HS - there is:
2. Identification of possible weak signals and sources of such signals;
3. Accessing and monitoring these sources (which may involve the creation of new data, e.g. via surveys), or the translation of raw data of some form into material useful as signals (e.g. turning press reports or social media postings into recognisable signals);
4. Assessing the potential significance of these signals (which may require feeding them directly into an automated alert system, but which will generally require some human judgement as to what, if any, action is undertaken in response to alerts);
5. Reporting on the nature and quality, and urgency in terms of decision-making and action, of the judgements based on these signals. The users of HS may then make decisions informed by the activity; and it may be standard practice or an additional decision to share the HS output with other parties.

## 3. AI for Horizon Scanning

Artificial Intelligence (AI) can be applied to each of these stages (and other computer-based tools may also be brought into play). The next few lines will consider some potential AI applications across the different stages. Given the rapid pace of advance of AI, AI-related, and AI-underpinning technologies, it is very likely that the ideas presented below will be superseded by developments in the near future; but they give a first indication of the range and scope of AI applications, many of which are already relatively easy to implement, assess, and learn from.

In particular, two types of AI may be relevant to HS tasks:

Machine Learning (ML) has been applied in many situations where it may be possible for computer systems to detect patterns in complex data, for example discriminating in visual images between human organs with cancerous and non-cancerous features, in audio signals between the voices of different people (and simulacra of individual voices can likewise be achieved); or in forecasting journey times or identifying some instances of fraud. ML can also be use to augment modelling of complex phenomena, such as the dynamics of disease transmission (Ali et al., 2022; Dairi et al., 2021; Fan et al., 2018; H. Purwins et al., 2019).

Chatbots are able to generate text in response to queries and commands, and users of tools such as ChatGPT or Claude can effectively have "conversations" (interchanges where the user prompts the chatbot with a query, and the chatbot responds with an answer, both using natural language – even speech recognition and synthesis in some cases). These draw on Large Language Models (LLMs) outlining likely connections between words (based on ML from, and analysis of, huge amounts of text available online), and working within rules of grammar and patterns of text identified as associated with specific areas of discussion (Dam et al., 2024; Kumamoto et al., 2023).

Each of these AI approaches may augment human activity, including augmentation of human use of more basic computer models and statistical analysis. For example, ML may rapidly detect anomalies in large data sets being employed in (non-AI) data analytics work (A. B. Nassif et al., 2021); chatbots can locate relevant text in large bodies of literature, and provide natural language interfaces with (non-AI) computer programs (A. Bhattacharjya et al., 2022). It should be borne in mind that AI systems may not provide perfect results and *need to be reviewed by humans*. This is especially important when AI outputs are being used to guide decision-making; often such outputs could be considered more as provocations, or starting points for further discussion and analysis, rather than as conclusions. AI may support an "early warning system", issuing alarms which draw human attention to a particular situation or change in circumstances (Kaur Sidhu, 2025; Villanueva-Miranda et al., 2025; WHO, 2021).

The following sections consider the various stages of HS. The main topic of interest is the emergence or re-emergence of Infectious Diseases (IDs), as seen mainly from a Public Health perspective. This is a topic of huge importance for human health and well-being, and one that requires both technical expertise and political wisdom for challenges – including signs of impending threats – to be effectively handled. While AI can support decision-making, it is important that its use does not undermine the technical training, learning from practical experience, and awareness of the wide spectrum of issues and of affected parties that are associated with Public Health policies in relation to such challenges. We are a long way from the point at which AI can be safely entrusted with making important decisions.

## 4. Establishing the Framework for Horizon Scanning

AI applications are relatively limited at this step in systematic HS, since this is essentially a matter of reflecting on organisational policy and strategy, setting goals and priorities, and other tasks requiring the value judgements (and, quite possibly, reflection on tensions between different values, some process of and mediation between different sets of interest). *Human judgement is inevitably central here*, and participation in discussions about the scope for, and scoping of, HS in this field can enhance learning about the range of issues involved – including economic, ethical, scientific and social dimensions of ID threats (Abdelhakim et al., 2024; Sutherland et al., 2012).

While not supplanting such discussions, AI tools can be used to provide some information that can help inform key decision makers and technical experts alike. AI outputs can be used to provoke or enhance discussion and reflection – rather than supplanting these activities (Drosos et al., 2025; Lee et al., 2025). For instance, AI-assisted search engines can retrieve and summarise information on relevant topics – for example, a chatbot might be asked questions such as: "What sorts of HS system are appropriate for addressing issues of Public Health, like threats posed by new or re-emerging IDs?" "IDs are being explored by national or international Public Health agencies?" "What publications are available on this topic?" "How can HS assist and augment surveillance efforts in the field of ID?" "Which academic or public sector organisations might have capabilities in HS for IDs, that would be valuable to co-operate with (for example, pooling resources to save redundant efforts)?"

The answers to such questions can be valuable starting points for expert discussion and review. They can contribute to the establishment of a mission statement for the organisation's HS efforts if it is decided to go ahead with these. Chatbots can also be employed to assist group discussions, for example by summarising statements concerning points of view, providing brief versions of lengthy documents, and the like (Labadze et al., 2023; Nordmann et al., 2025).

## 4.1 On Frameworks for HS

Systematic HS will typically be undertaken on behalf of some organisation, in order to help it attain its objectives. At a broad strategic management level, this may be a matter of informing plans for the organisation, for example considering whether its resources, capabilities, missions, and operating environment are liable to change in substantial ways in the near, medium, or longer-terms. The meaning of these time horizons will vary across different types of organisation – change is typically much more rapid in some fields and locations than in others, though even the longest-established bodies may be subject to sudden "shocks" or upheavals that may emerge slowly but then accelerate (P. Hines et al., 2019; Miles et al., 2016). HS is often conducted on behalf of some part of a large organisation, however: it may concern specific areas or types of operation, it may involve internal or external features of the organisation (or both). Importantly, the outputs of HS may be of relevance to stakeholders other than those in the organisation, or part of the organisation, concerned. These may be, for example, sponsors, partners, customers and suppliers, employees and communities, the general public. HS information may be shared partially or comprehensively, to few or many stakeholders (Boult et al., 2018; A. Hines et al., 2018; Wintle et al., 2020).

HS will often focus on known risks, which will often already be subject to some surveillance and preparedness efforts (Georghiou & Cassingena Harper, 2013; Loveridge & Saritas, 2012). However, there may also be interest in "wild cards" – less probable phenomena, but ones with high impact if they do manifest (Bengston, 2013; Bishop, 2009; Miles et al., 2016). The precise functions for which HS is intended will thus determine a great deal about the activity, including the resources to be dedicated to it; whom it reports to; whether channels are established for urgent reporting to be instituted alongside more regular reports (indeed, some kinds of reporting may be available on a continual basis, in the form of webpages updated in real time); what form reports should take. Furthermore, the focus of the HS, in terms of whether it covers a wide or narrow range of themes, is a major issue (Boult et al., 2018; P. Hines et al., 2019).

In the case of HS for Emerging or Re-emerging Infectious Diseases, as is the focus of VigIA, aiming to appraise the feasibility of an AI-assisted HS system, what sorts of things might this consider in terms of HS? HS might well not just be aimed at generating knowledge concerning the nature of a possible ID outbreak, but also concerning a wide range of phenomena that might shape the likelihood and nature of such an outbreak. These phenomena might constitute "weak signals" of such likelihood and nature of an outbreak, and the threat that an initial outbreak could develop into a pandemic. These could include developments "over the horizon" related to:

- Possible sources of such IDs (zoonotic sources – including those from xenotransplantation as well as the more familiar contact between humans and animals, ancient viruses emerging from melting permafrost, lab leaks, and so on). The main topics here are probably the matters of (a) whether there are trends in the likelihood of such sources increasing in prevalence or proximity of human contact; (b) the reporting of new or unusual disease outbreaks in wild or farmed species of interest (e.g. bird flu).
- The features of IDs – possible types of disease agent (bacterial, viral, protozoa, fungal, prions, and possibly new emerging types related to "obelisks", etc.) and modes of transmission (airborne, STIs, other vectors) of these agents, disease characteristics such as lethality and morbidity, nonsymptomatic transmission, etc.;
- Social arrangements that would foster disease transmission (cf. "modes of transmission" above). Population density in urban areas, amount of long-distance travel, and significant breakdown of social order (war, civil unrest, etc.) that might impact not only health provisions, but also nutrition, water supplies, sanitation, hygiene, and the like.
- Bioscience developments, for example capabilities for rapid detection and identification of the nature of an ID, and the carriers of such a disease in a population.
- Pandemic preparedness, such as investment in, and maintenance of, facilities and equipment that allow for treatment of patients and protection of health (and other workers (e.g. Personal Protective Equipment such as masks, air purification systems, etc.); and Medical advances, such as the capacity to determine effective treatment of IDs, and the availability of such treatments.
- Sociopolitical circumstances, such as government support for ID Surveillance and for Public Health measures; and, among the population, vaccine hesitancy and levels of trust in medical and other expertise.

Clearly, the ability to gain early warning of epidemics and pandemics would be extremely valuable. But data concerning the entire panoply of issues outlined above provides information, for decision makers, about topics that are liable to prove relevant for preparedness and risk management, for Public Health and R&D strategies, and for communication with stakeholders.

As noted above, AI's role in supporting the development and implementation of a framework for systematic HS is mainly limited to its capabilities for locating, identifying, and summarising available information about such frameworks. This information can be helpful for decision-makers – and the power of chatbots in these respects may make quite a persuasive case for further exploration of AI systems in HS for ID. However, the current state of the art is such *that human expertise will be required*: (a) to check for and weed out errors and "hallucinations" (fabrications) in the material produced by chatbots, and (b) to tailor the information to the needs, resources, and organisational structures that are in play.

## 4.2 Weak Signals and their sources

"Weak Signals" (WSs) are generally described as being rather ambiguous items of information, that can easily be interpreted in various ways. Most data sets are "noisy", with errors and random variations meaning that a single anomalous item of data is rarely regarded as sufficient to cause an alarm. When the anomaly is a striking one, suspicions may be raised as to the validity of the data. Supplementary evidence is sought to check whether we are dealing with "noise" rather than a signal. But such evidence may itself be ambiguous, and it may require a mounting body of evidence to clarify matters. This is often the case with social trends, for example (Bengston et al., 2024; Miles et al., 2016; Saritas & Smith, 2011; Zhao et al., 2023).

WS will vary across topics. For example, devotion of R&D efforts to a given field, the emergence of patents in this field, clinical trials – all of these may indicate that major technological developments are likely in this field. Where it concerns tools for disease detection and identification, for new types of vaccine and vaccine development methods, new treatments for diseases and new ways of delivering these treatments, and so on, WS can be important for HS focused on medical approaches (P. Hines et al., 2019; Miles & Saritas, 2012; Urquhart & Saunders, 2017). Such HS can inform research efforts, training, and other aspects of forward planning. In contrast, HS directly focused on the threat of new and re-emerging IDs might consider WS associated with (a) the origins and initial outbreak of the ID in a human population and (b) the spread of the ID beyond the initial affected population to a much wider one (Petrokofsky et al., 2025; Villanueva-Miranda et al., 2025). Issues such as disease lethality, transmissibility, incidence and severity across different social groups, and the like, would also be of interest. While predicting specific outbreaks may be difficult, factors conducive to zoonotic transmission of disease can be identified - for instance, presence of ID in animal hosts, levels of human interaction with such hosts, environmental changes associated with climate change and/or intrusion into habitats. When it comes to the spread of an ID, issues such as mobility, population health, population density, social breakdown associated with war or other conflict, and the like may be of more interest (Lin et al., 2025; Marmot et al., 2010; UKHSA, 2025a) .

Some of these factors are ones that change slowly, others can change much more rapidly. In either case, WS may be extracted from publications as well as from more immediate news sources and social media. Chatbots then can be used *to supplement expert views* about the potential WS that might provide useful information about ID outbreaks and dynamics; and, of course, they may be interrogated as to potential WS related to such other topics as developments in medical technology and treatments, in Public Health policies and problems, and so on. As always with chatbot use, it can be fruitful to explore the consequences of posing questions (prompts) in various ways – and this might involve requesting that the chatbot respond by providing its estimate of what particular types of observer might think.

By way of illustrating the strengths and limitations of chatbots as tools to assist thinking about HS topics, consider the results of a query to Copilot on 08/04/2024. This asked a follow-up question to the initial prompt "What is the range of weak signals that can provide early warning of disease outbreaks?", that follow-up being "Can you talk of different types of weak signals?" – Answers in Box 1. A further query, "Can you outline weak signals that might predict the spread of disease beyond the initial outbreak?" elicited a practically identical answer.

Box 1 – Copilot on Types of Weak Signal

> Certainly! **Weak signals** are subtle indicators or early signs that might not be immediately obvious but can provide valuable insights. Let's explore some types of weak signals:
> 1. Syndromic Surveillance:
>    - This involves monitoring **symptoms and health data** from various sources to detect outbreaks or unusual patterns.
>    - For example, an increase in **fever-related visits** to emergency departments could signal an emerging infectious disease outbreak.
> 2. Social Media Trends:
>    - Analysing social media platforms can reveal weak signals related to Public Health.
>    - **Hashtags**, mentions, or keywords related to symptoms or diseases might indicate a potential outbreak.
> 3. Environmental Changes:
>    - Unusual changes in the environment can serve as weak signals.
>    - For instance, a sudden increase in **mosquito activity** might hint at a potential **vector-borne disease** outbreak.
> 4. Animal Behavior:
>    - Abnormal behavior in animals can be an early warning.
>    - Bird die-offs, mass strandings of marine mammals, or unusual migration patterns could signal environmental or zoonotic threats.
> 5. Healthcare Worker Reports:
>    - Frontline healthcare workers often notice subtle changes.
>    - Their observations, such as an increase in specific symptoms, can be valuable weak signals.
> 6. Pharmaceutical Sales Data:
>    - Unusual patterns in drug sales can indicate health issues.
>    - For example, a sudden surge in **antiviral medication** sales might suggest an emerging outbreak.
> 7. School Absenteeism:
>    - A rise in student absenteeism due to illness can be an early indicator.
>    - Schools can serve as sentinel sites for monitoring community health.
> 8. Search Engine Queries:
>    - Analysing search engine trends can reveal weak signals.
>    - An increase in searches related to specific symptoms or diseases might signal an outbreak.
>
> Remember, detecting weak signals requires vigilance and collaboration across various sectors. Combining multiple sources of information enhances our ability to respond swiftly to potential health threats.

## 4.3 Accessing and monitoring data sources

The availability of data in electronic form has increased tremendously over recent decades. The scope for processing qualitative material, especially natural language text or speech, rather than quantified statistics; but also graphic images as provided by handheld cameras as well as those carried by remote sensing satellites has also expanded, and this is one way in which data "capture" can be enhanced by application of AI. The notion of big data emerged as a way of describing the large and complex, and often rapidly growing, data provided by many information systems. Big data challenges traditional statistical analysis but is amenable to processing by AI systems (Kumar et al., 2024; Maslej et al., 2025; Sapkota et al., 2025).

Much experimentation is underway in applying these capabilities to health-related topics, where a vast amount of data is generated routinely by modern health and medical systems, and where there is almost certainly much scope for making use of social, economic, geographical and even meteorological information that could impinge upon human well-being (Baloch et al., 2023; Resch et al., 2025). While many types of data are currently dispersed and patchily accessible and are often made available only slowly and with limitations in terms of format and metadata (Saberi et al., 2025), the likelihood is a combination of AI and changes in organisational routines will be reduce these problems.

## Data Sources

Much information is posted online by the WHO and various national agencies, though often there are delays between data capture and publication (Felix, 2024; WHO, 2025). Most of these delays are simply a matter of fact-checking and slow bureaucratic processes, but sometimes there are also political interventions that hinder rapid reporting (GAO, 2022; Morabia, 2025).

Strong signals that an ID outbreak is underway are evident in Health Surveillance data, such as the "dashboards" produced by the CDC in the USA and the HSA in the UK, that chart recorded cases of specific IDs in these countries (CDC, 2025; UKHSA, 2025b). Weak signals, however, may supplement or precede such information. Newspaper and other news sources may be an early source of information on outbreaks, often reproducing Public Health surveillance reports, but in some cases reflecting other origins (Notably, the earliest reports of COVID-19 came from individual doctors in Wuhan, at a point when the authorities were, for whatever reason, failing to acknowledge the growing problem). News media, then, may present a mix of strong and weak signals; and it may be difficult to differentiate between the two. News reports are typically unscientific and expressed in natural languages, and there may be multiple reports of the same instance. This has led the UKHSA (Health Security Agency) to employ LLMs to scan news reports, avoid duplicate reports, and summarise what type of ID or symptom is the subject of reports, where and when. The source of news reports is GDELT, which provides information derived from broadcast, print, and web news sources around the world, in over 100 languages (Harris et al., 2024).

With much of the world now having access to smartphones and the internet, there is growing scope for relevant information to be posted online, in reports from news outlets or organisations of other types, in blogs and social media. Scientific and medical online "communities" and chatgroups are likely to be important venues for early discussion of emerging threats to Public Health, as noted above, questions posed to search engines (and more recently to AI-enhanced query systems). Such sources can be scanned for references to symptoms and clusters of symptoms of disease among humans.

Additionally, such sources of information can be scanned for references about disease in animal populations (since many IDs are zoonotic in origin). Chatgroups featuring farmers, veterinarians and agronomists might be fruitful venues to explore. Various online sources can also be scanned for information about the evolution of conditions that might make zoonotic transfer of disease more likely (e.g. increased intrusion of human activities into natural environments) or which might make it harder to contain disease outbreaks (war or other breakdowns of civil order, etc). Different WS may be relevant to the likelihood of origin of a new or re-emerging ID, and to the spread of an ID once an initial outbreak is apparent.

## From Traditional to Digital Data Access

Traditionally, HS would often be undertaken with a team of research assistants trawling through large volumes of printed material. Back in 2018, a blog post from the UKHSA reported that this organisation had setup an HS team, who spent half of each day "combing nearly a hundred sources on the internet for rumours of diseases and "unusual incidents" around the world (Lloyd, 2018):

> (…) *"We look at everything – official reports from international organisations like the WHO and Ministries of Health, international and local media, and even Facebook and Twitter, because the majority of the world's first news about infectious disease events now comes from unofficial sources, including newspapers, social media and other internet resource (…) One of the main things we look for are rumours; rumours from local or social media of undiagnosed diseases, unusual events, haemorrhagic fevers (serious viral infections characterised by sudden onset of fever and bleeding) and/or large numbers of deaths. When we find something, we record it in our database and then go to work searching other sources for more information."*

As noted above, the UKHSA has now been experimenting with using chatbots and GDELT news sources for related purposes, where the raw data is translated into material useful as signals – in this case, turning press reports and social media postings into recognisable signals.

With the explosive increase of online publication, AI-assisted web crawling and scraping tools can scan huge volumes of data to extract relevant WS information, and via chatbot features to present results in readily comprehensible forms. To summarise, there is considerable scope for applying such AI tools to:

- Locating relevant information sources (e.g. new discussion groups that may touch on matters of interest, social media queries, etc.);
- Scraping meaningful text from these sources and, where necessary, translating into a common language – and creation of text from speech recordings can be effected through AI;
- "Cleaning" data by eliminating duplications, correcting typographical and similar errors;
- Extracting key information (e.g. on reported symptoms, diagnoses, mortality);
- Presenting results in forms that are easily comprehensible.

Such AI applications can be used for HS concerning ID emergence and spread. But they can also be applied to other matters of concern, such as development of disease detection, identification, prevention and treatment techniques. Here we may look to other specialist databases – for example, databases of patents or scholarly publications. The latter have been mined extensively by researchers in recent years, looking, for example, at the numbers of patents being taken out of papers being authored in specific fields, the frequency with which patents/papers are cited by later patents/papers, and so on (Krestel et al., 2021).

## AI and Data Creation

As if the amount of available data were not itself impressive, there is also scope for the creation of new data, e.g. expert or practitioner surveys, where AI can support survey design, administration and analysis. For example, Delphi Survey questions can be inspired by those used in earlier survey or translated across languages. Likewise for response options (Hasson et al., 2000; Teig et al., 2023).

Chatbots can be queried as to what they impute expert responses to Delphi or other questions might be. It is readily possible to request a chatbot to provide responses that a panel comprising a particular type, or variety of types, of experts might respond to survey queries, and in the course of VigIA studies we have experimented with requesting estimates of the likely distribution of responses from a panel of 100 experts concerning, for example, the likely next occurrence of a pandemic, and the most likely disease agents that might be involved in this. These answers should not be naively interpreted as the responses of an omniscient computer to all the available data on these topics – they are merely hypothetical accounts, based on analyses of what has been said in the past about such topics. The chatbot could further be prompted, too, to present views where the panel responses exhibit high levels of disagreement, for example.

Nevertheless, such "fabricated data" may be of some interest – for instance leading experts to review why it might be that the chatbot response is to give bacterial disease agents many more "votes" as being implicated in future pandemics than did the real-life panel consulted in the VigIA project, where the Brazilian experts, considering the likely disease agent in future ID outbreaks, were overwhelmingly in favour of this agent being a virus. However, a chatbot, asked to say what the distribution of views of potential threats would be in an expert group, suggested that a substantial minority of the experts would have considered bacterial IDs to be a real possibility – Figure 1. The reasons for this divergence, and whether the chatbot judgement should inspire further reflection, is something that could be fruitfully improved in a workshop, not least as a way of offsetting "groupthink", which has been blamed for some failures of responses to the COVID-19 pandemic (Forsyth, 2020).

Figure 1 – Delphi x Copilot responses about agent of future pandemics (June/2023)

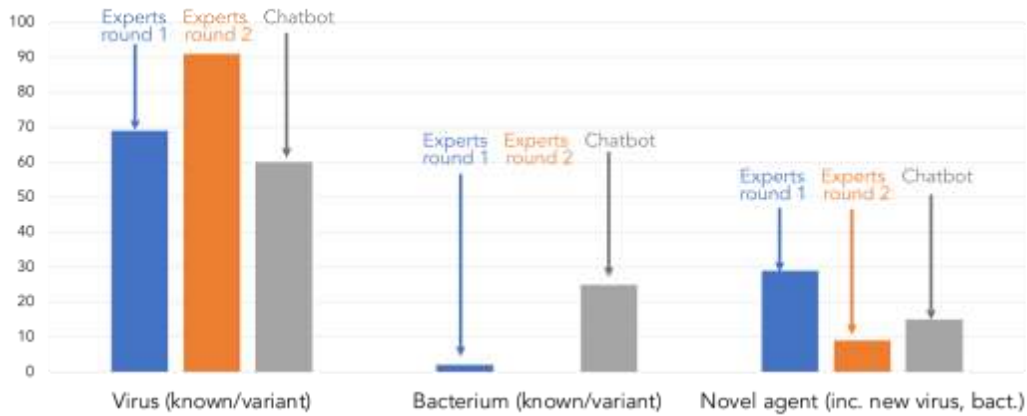

Source: VigIA project

## The Significance of Signals

The identification and interpretation of signals are central to Horizon Scanning. Some signals are strong ones, for example, data on recorded cases of a new ID can tell us about the presence of this disease in a population, whether it is becoming more or less prevalent, and so on (A. Hines et al., 2006; Urquhart & Saunders, 2017). The entry of a new medicine into clinical trials can tell us that important stakeholders believe that there is a reasonable prospect of new treatments becoming widely available in months or a few years time. Even data on research funding is indicative of expectations that new knowledge can be generated in a field that will feed into future applications, though often knowledge results in unanticipated applications.

Weak Signals, in turn, are by nature ambiguous – they are weak precisely because their meaning is a matter of human interpretation, where assumptions about links between the observed data and inferences about possible (future) developments that it may relate to are inherently uncertain. A WS may be an early indicator of an emerging trend, but this will only be known for sure once the trend has or has not manifested. If the WS was an anomaly or unusual phenomenon of some kind, too, it may be of significance other than that which was initially anticipated. On the other hand, a WS may just be "noise" (a more-or-less random fluctuation in data, or one caused by deficiencies in the signal detection apparatus), or a "blip" (a real, but very temporary, state of affairs that has very little significance for the overall situation). At one extreme, the experts may discount a WS entirely, as being unbelievable or entirely trivial: the data source may be seen as unreliable, or the apparent change as something that can be readily discounted (Ansoff, 1975; Eulaerts et al., 2025; Miles et al., 2016; Zhao et al., 2023).

A WS will typically be interpreted as possibly indicative of some change in state of part of the system that concerns us, whether this be the existence and prevalence of novel IDs, medical technologies, or social and environmental circumstances that may impact Public Health. If this could potentially imply a large change in state, and/or that part of the system is critically important, then there is reason to pay attention to the WS – looking (or waiting) to see if other data are consistent with this. Experts may regard the data as something worthy of attention, or of more active further exploration; they may even see it as a cause for alarm (or more positive excitement). Having judged how far the signal is more than just noise or a blip, the issue is how significant this *might* be for assessment of the element(s) of the system it pertains to; and how important these parts of the system are.

The importance of the factor in question will be judged on the basis of a model of some sort, in which different elements of the system are related together. This model could be an implicit, purely mental model, held by those examining the topic. It could be a more formal model, where the mental model is made explicit by some process of cognitive mapping, or a statistical model or a simulation model; the latter take us into the realm of "predictive AI", where the model is at least in part generated by an AI system.

Traditionally, the significance of WS has been a matter of expert judgement (Ansoff, 1975). This may rely upon one particular expert, or be determined through teamwork (usually in the course of group discussions, but a survey technique such as Delphi could be involved to elicit views on which of a set of WS is most likely to be important, and in which way. It would be possible to supplement such human judgement with enquiries made of a chatbot. For example, when CoPilot was asked "Recently, there have been reports of avian influenza being responsible for conjunctivitis in cattle. Is this a weak signal of implications for human Public Health? If so, what are the implications?". The response did not directly answer the question "is this a WS?" but clearly proceeded to interpret it as one, and then provided suggestions as to possible risks, requirements for further information, and preventative measures. Further chatbot responses could be elicited to address each of these dimensions of the answer that has been given.

The response to the question about bovine conjunctivitis suggests that it is clearly seen as signal, and thus did not directly answer the question "is this a WS?", a little further exploration can feature a more ambiguous signal. CoPilot responses to a subsequent question "Would an upsurge in search engine enquiries about conjunctivitis be a weak signal of implications for human Public Health? If so, what are the implications?"

In this case, alternative explanations for the phenomenon are first outlined, which is a helpful reminder to be cautious in treatment of such data. The chatbot suggests requirements for further monitoring, suggests Public Health responses and preventative measures, and outlines potential Public Health impacts associated with an ID outbreak being underway.

This rather simple trial of just one of the many chatbots that are now available, on some fairly rudimentary WS-type inputs, confirms that these tools may be helpful for accompanying human expert interpretation of such data. A "trained" chatbot could be linked via APIs to a search engine or source of news data, and report when WS are being observed, and on their possible implications. Clearly, it would not be appropriate for alarm warnings to be sounded for every single new observation of potential significance, so the tool would need to be engineered to respond in a proportionate fashion. This will mean taking into account the strength of the signal (and the quality of the data source) and the risk level involved (where risk is often seen as comprising both likelihood of the phenomenon, and the severity of impact if it does occur). The output can also be rendered in different ways – a "dashboard" could be generated with the nature and severity of the threat, and possibly also key further observations and preparedness for action, represented by different graphical features. Such a dashboard would be used to alert and provoke human expertise in WS interpretation.

As in other instances, chatbots may support Expert Group workshops, which may be held to discuss the nature and meaning of WS. They can support improved interfaces between conventional computer simulation models and their users. For example, the inputs to, and outputs from, such models may involve natural language interfaces, and outputs, also, could be rendered in a variety of ways, for a variety of users, including graphic and similar representations.

## Risks and Risk Management

The use of Artificial Intelligence (AI) in Horizon Scanning (HS) for Infectious Diseases (IDs) presents significant opportunities, but also introduces risks and limitations that require careful management. The primary challenge lies in the inherently imperfect nature of AI systems, which may include:

- **Errors and "Hallucinations" (Fabrications)**: AI tools, such as chatbots, can produce material that contains errors and "hallucinations". Risk management requires that human expertise will be required to check for and weed out these fabrications.
- **Overreliance on AI**: Excessive dependence on automated systems may diminish human judgment, especially in complex scenarios requiring ethical, social, and political considerations. Thus, implementing strategies that mandate and enhance human expertise at critical stages, ensuring AI serves as a *support tool* rather than a replacement for human judgment.
- **Data Quality and Bias**: AI effectiveness is contingent on the quality of input data. Incomplete, biased, or manipulated data can distort outcomes and compromise analyses. This requires diversifying input sources (e.g., news, social media, official reports) to reduce single-source bias, applying AI tools for data cleaning and standardisation, and mandating human expert review to validate outputs.

- **Privacy and Security**: The use of sensitive data, such as Public Health records or social media content, raises ethical and legal concerns regarding privacy and data protection that demand human-led governance to establish ethical frameworks, using AI primarily to extract anonymised data from sources like social media, and ensuring legal compliance to maintain the public trust for effective Public Health measures.
- **The Act of Decision-Making**: Whereas AI can support decision-making, it is a long way from the point at which AI can be safely entrusted with making important decisions. AI outputs should be considered more as *provocations or starting points* for further discussion and analysis, rather than as conclusions. Human review is especially important when AI outputs are being used to guide decision-making.
- **Undermining Human Expertise**: It is vital that the use of AI does not undermine the technical training, learning from practical experience, and awareness of the wide spectrum of issues and affected parties associated with Public Health policies.
- **Ambiguity of Weak Signals (WS)**: WS are, by nature, ambiguous, and their identification and meaning is a matter of human interpretation. AI can assist interpretation, but the risk that a signal may just be a "noise" must be managed with further monitoring and the assessment of risk level (likelihood and severity of impact).

## Conclusions and Recommendations

This review explored the integration of Artificial Intelligence (AI) into Horizon Scanning (HS) processes, with a focus on identifying and responding to emerging threats and opportunities in the field of Infectious Diseases (IDs), drawing insights from the VigIA project. The findings contribute to the growing body of Foresight literature by demonstrating the **potential and limitations of AI** in Public Health preparedness.

### Key Conclusions

- **Potential for HS Enhancement**: AI—including Machine Learning (ML) and generative models (such as chatbots and Large Language Models - LLMs) — can significantly enhance the HS process. This includes improvements in signal detection, data monitoring, scenario analysis, and decision support.
- **Efficiency in Data Capture**: AI proves crucial for handling "big data" in health. AI-assisted web crawling and scraping tools can scan huge volumes of data to extract relevant WS information including from news (e.g. via GDELT), social media, and specialised databases. These tools can "clean" data and extract key information to create recognisable signals.
- **Support Role for Discussion and Reflection**: AI's role in establishing the HS framework is mainly limited to locating and summarising available information. Chatbots can serve as provocations or starting points for expert discussion, helping to inform key decision-makers and technical experts alike and offsetting "groupthink".

## Recommendations

- **Implementation of Early Warning Systems**: AI should be used to support an "early warning system", issuing alarms which draw human attention to a particular situation or change in circumstances. It is recommended to develop dashboards that are used to alert and provoke human expertise in WS interpretation, representing the nature and severity of the threat graphically.
- **Mandatory Human Review**: Given the imperfect nature of AI, human judgement is inevitably central in all stages of HS, especially in assessing the significance of signals and in decision-making. Chatbot outputs must be tailored to the needs, resources, and organizational structures that are in play.
- **Continuous Exploration of Chatbots**: The capability of chatbots to provide hypothetical accounts of expert responses ("fabricated data") should be used as a fruitful tool for further reflection, allowing experts to review why the chatbot's judgment might diverge from their own.
- **Development of Enhanced Interfaces**: It is important to develop natural language interfaces for conventional computer simulation models and other computer programs, improving the user experience for both inputs and outputs from modeling.

# References

A. B. Nassif, M. A. Talib, Q. Nasir, & F. M. Dakalbab. (2021). Machine Learning for Anomaly Detection: A Systematic Review. *IEEE Access*, *9*, 78658–78700. https://doi.org/10.1109/ACCESS.2021.3083060

A. Bhattacharjya, A. Punyamurthi, S. Atota Lakshmi, A. Iragala, M. C. Thallapureddy, & S. K. Kannaiah. (2022). An Exploratory Study on Chatbots. *2022 2nd International Conference on Electronic and Electrical Engineering and Intelligent System (ICE3IS)*, 340–344. https://doi.org/10.1109/ICE3IS56585.2022.10010154

Abdelhakim, D., Stevance, A.-S., Bridgewater, P., Sullivan, L., Newlands, M., Barrow, S., Rengel-Gonçalves, A., Datta, R., Luga, W., Lazurko, A., Pintér, L., Schweizer, V., Gluckman, P., Sridhar, H., Saner, M., Hinwood, A., Jabbour, J., Gutiérrez-Espeleta, E. E., & Mangalagiu, D. (2024). *A guide to anticipation: Working Paper on Tools and Methods of Horizon Scanning and Foresight*. International Science Council. https://doi.org/10.24948/2024.10

Ali, A., Abd Razak, S., Othman, S. H., Eisa, T. A., Al-Dhaqm, A., Nasser, M., Elhassan, T., Elshafie, H., & Saif, A. (2022). Financial Fraud Detection Based on Machine Learning: A Systematic Literature Review. *Applied Sciences*, *12*(19). https://doi.org/10.3390/app12199637

Amanatidou, E., Butter, M., Carabias, V., Könnölä, T., Leis, M., Saritas, O., Schaper-Rinkel, P., & van Rij, V. (2012). On concepts and methods in horizon scanning: Lessons from initiating policy dialogues on emerging issues. *Science and Public Policy*, *39*(2), 208–221. https://doi.org/10.1093/scipol/scs017

Ansoff, H. I. (1975). Managing Strategic Surprise by Response to Weak Signals. *California Management Review*, *18*(2), 21–33. https://doi.org/10.2307/41164635

Baloch, L., Bazai, S. U., Marjan, S., Aftab, F., Aslam, S., Neo, T. K., & Amphawan, A. (2023). A review of big data trends and challenges in healthcare. *International Journal of Technology*, *14*(6), 1320–1333.

Bengston, D. N. (2013). *Horizon scanning for environmental foresight: A review of issues and approaches*. U.S. Department of Agriculture, Forest Service, Northern Research Station. https://doi.org/10.2737/nrs-gtr-121

Bengston, D. N., Mauno, T., & Hujala, T. (2024). Horizon Scanning: A Process for Identifying Emerging Signals of Change Shaping the Future of Natural Resources Management. *Society & Natural Resources*, *37*(7), 1108–1115. https://doi.org/10.1080/08941920.2024.2335392

Bishop, P. (2009). Horizon scanning why is it so hard. *Houston: University of Houston. Retrieved at February*, *20*, 2016.

Boult, M., Jassal, V., de Souza, R. S., Byatt, G., Kirkman-Page, J., Smith, K., Falarin, S., Massey, M., Thompson, C., & Falconer, S. (2018). *Horizon Scanning: A Practitioner's Guide*.

Brisse, M., Vrba, S. M., Kirk, N., Liang, Y., & Ly, H. (2020). Emerging concepts and technologies in vaccine development. *Frontiers in Immunology*, *11*, 583077.

CDC. (2025). *NNDSS Dashboards Monitor Data Quality*. National Notifiable Diseases Surveillance System (NNDSS). https://www.cdc.gov/nndss/in-action/dashboards.html

Cuhls, K. E. (2020). Horizon Scanning in Foresight – Why Horizon Scanning is only a part of the game. *Futures & Foresight Science*, *2*(1), e23. https://doi.org/10.1002/ffo2.23

Dairi, A., Harrou, F., Zeroual, A., Hittawe, M. M., & Sun, Y. (2021). Comparative study of machine learning methods for COVID-19 transmission forecasting. *Journal of Biomedical Informatics*, *118*, 103791.

Dam, S. K., Hong, C. S., Qiao, Y., & Zhang, C. (2024, June 17). *A Complete Survey on LLM-based AI Chatbots*. arXiv.Org. https://arxiv.org/abs/2406.16937v2

Drosos, I., Sarkar, A., Xiaotong, Xu, & Toronto, N. (2025, January 28). *'It makes you think': Provocations Help Restore Critical Thinking to AI-Assisted Knowledge Work*. arXiv.Org. https://arxiv.org/abs/2501.17247v1

Eulaerts, O., Grabowska, M., & Bergamini, M. (2025). *Weak signals in Science and Technologies—2024. KJ-01-25-060-EN-N (online)*. https://doi.org/10.2760/6571994%2520(online)

Fan, S.-K. S., Su, C.-J., Nien, H.-T., Tsai, P.-F., & Cheng, C.-Y. (2018). Using machine learning and big data approaches to predict travel time based on historical and real-time data from Taiwan electronic toll collection. *Soft Computing*, *22*(17), 5707–5718. https://doi.org/10.1007/s00500-017-2610-y

Felix, S. E. B. (2024). A standard framework for evaluating large health care data and related resources. *MMWR Supplements*, *73*.

Forsyth, D. R. (2020). Group-level resistance to health mandates during the COVID-19 pandemic: A groupthink approach. *Group Dynamics: Theory, Research, and Practice*, *24*(3), 139.

GAO. (2022). *Scientific Integrity: HHS Agencies Need to Develop Procedures and Train Staff on Reporting and Addressing Political Interference*.

Georghiou, L., & Cassingena Harper, J. (2013). Rising to the challenges—Reflections on Future-oriented Technology Analysis. *Future-Oriented Technology Analysis*, *80*(3), 467–470. https://doi.org/10.1016/j.techfore.2012.10.009

Georghiou, L., Harper, J. C., Keenan, M. P., Miles, I., & Popper, R. (Eds). (2008). *The Handbook of Technology Foresight. Concepts and practice*. Edward Edgar Publishing Limited.

Grassly, N. C., & Fraser, C. (2008). Mathematical models of infectious disease transmission. *Nature Reviews Microbiology*, *6*(6), 477–487. https://doi.org/10.1038/nrmicro1845

H. Purwins, B. Li, T. Virtanen, J. Schlüter, S. -Y. Chang, & T. Sainath. (2019). Deep Learning for Audio Signal Processing. *IEEE Journal of Selected Topics in Signal Processing*, *13*(2), 206–219. https://doi.org/10.1109/JSTSP.2019.2908700

Harris, J., Laurence, T., Loman, L., Grayson, F., Nonnenmacher, T., Long, H., WalsGriffith, L., Douglas, A., Fountain, H., & Georgiou, S. (2024). Evaluating large language models for public health classification and extraction tasks. *arXiv Preprint arXiv:2405.14766*.

Hasson, F., Keeney, S., & McKenna, H. (2000). Research guidelines for the Delphi survey technique. *Journal of Advanced Nursing*, *32*(4), 1008–1015.

Hines, A., Bengston, D. N., Dockry, M. J., & Cowart, A. (2018). Setting Up a Horizon Scanning System: A U.S. Federal Agency Example. *World Futures Review*, *10*(2), 136–151. https://doi.org/10.1177/1946756717749613

Hines, A., Bishop, P. J., & Slaughter, R. A. (2006). *Thinking about the future: Guidelines for strategic foresight*. Social Technologies Washington, DC.

Hines, P., Yu, L. H., Guy, R. H., Brand, A., & Papaluca-Amati, M. (2019). Scanning the horizon: A systematic literature review of methodologies. *BMJ Open*, *9*(5), e026764.

Kaur Sidhu, B. (2025). Hallucinations in Artificial Intelligence: Origins, Detection, and Mitigation. *International Journal of Science and Research (IJSR)*, *14*(1), 8–15. https://doi.org/10.21275/SR241229170309

Krestel, R., Chikkamath, R., Hewel, C., & Risch, J. (2021). A survey on deep learning for patent analysis. *World Patent Information*, *65*, 102035. https://doi.org/10.1016/j.wpi.2021.102035

Kucharski, A. J., Russell, T. W., Diamond, C., Liu, Y., Edmunds, J., Funk, S., Eggo, R. M., Sun, F., Jit, M., & Munday, J. D. (2020). Early dynamics of transmission and control of COVID-19: A mathematical modelling study. *The Lancet Infectious Diseases*, *20*(5), 553–558.

Kumamoto, T., Yoshida, Y., & Fujima, H. (2023, December 8). *Evaluating Large Language Models in Ransomware Negotiation: A Comparative Analysis of ChatGPT and Claude*. https://doi.org/10.21203/rs.3.rs-3719038/v1

Kumar, Y., Marchena, J., Awlla, A. H., Li, J. J., & Abdalla, H. B. (2024). The AI-Powered Evolution of Big Data. *Applied Sciences*, *14*(22). https://doi.org/10.3390/app142210176

Labadze, L., Grigolia, M., & Machaidze, L. (2023). Role of AI chatbots in education: Systematic literature review. *International Journal of Educational Technology in Higher Education*, *20*(1), 56.

Lee, H.-P. (Hank), Sarkar, A., Tankelevitch, L., Drosos, I., Rintel, S., Banks, R., & Wilson, N. (2025). The Impact of Generative AI on Critical Thinking: Self-Reported Reductions in Cognitive Effort and Confidence Effects From a Survey of Knowledge Workers. *Proceedings of the 2025 CHI Conference on Human Factors in Computing Systems*, 1–22. https://doi.org/10.1145/3706598.3713778

Lin, L., Guo, X., Wu, Y., Sun, Y., Liu, D., Luo, Y., Xiang, Q., Li, T., Li, G., & Yu, W. (2025). Global Status of Infectious Diseases from January to June of 2025. *Zoonoses*, *5*(1), 967.

Lloyd, J. (2018, February 21). *Disease Detectives: Keeping track of new and emerging infectious diseases – UK Health Security Agency*. https://ukhsa.blog.gov.uk/2018/02/21/disease-detectives-keeping-track-of-new-and-emerging-infectious-diseases/

Loveridge, D., & Saritas, O. (2012). Ignorance and uncertainty: Influences on future-oriented technology analysis. *Technology Analysis & Strategic Management*, *24*(8), 753–767. https://doi.org/10.1080/09537325.2012.715477

Marmot, M., Atkinson, T., Bell, J., Black, C., Broadfoot, P., Cumberlege, J., Diamond, I., Gilmore, I., Ham, C., Meacher, M., & Mulgan, G. (2010). *Fair society, healthy lives—The Marmot Review: Strategic Review of Health Inequalities in England post-2010*. University College London.

Maslej, N., Fattorini, L., Perrault, R., Gil, Y., Parli, V., Kariuki, N., Capstick, E., Reuel, A., Brynjolfsson, E., & Etchemendy, J. (2025). Artificial intelligence index report 2025. *arXiv Preprint arXiv:2504.07139*.

Miles, I., & Saritas, O. (2012). The depth of the horizon: Searching, scanning and widening horizons. *Foresight*, *14*(6), 530–545. https://doi.org/10.1108/14636681211284953

Miles, I., Saritas, O., & Sokolov, A. (2016). *Foresight for Science, Technology and Innovation*. Springer.

Morabia, A. (2025). AJPH and the Threat of Political Interference in Scientific Publishing. *American Journal of Public Health*, *115*(5), 626–627. https://doi.org/10.2105/AJPH.2025.308100

Nordmann, K., Sauter, S., Stein, M., Aigner, J., Redlich, M.-C., Schaller, M., & Fischer, F. (2025). Evaluating the performance of artificial intelligence in summarizing pre-coded text to support evidence synthesis: A comparison between chatbots and humans. *BMC Medical Research Methodology*, *25*(1), 1–14. https://doi.org/10.1186/s12874-025-02532-2

Petrokofsky, G., Harvey, W. J., Petrokofsky, L., & Savilaakso, S. (2025). One Health Horizon Scanning: Contribution to Road Map. *Juno Reports*, *2*(1), 0003.

Ravetz, j., Miles, I., & Popper, R. (2011). *iKnow ERA Toolkit—Applications of Wild Cards and Weak Signals to the Grand Challenges & Thematic Priorities of the European Research Area*. http://community.iknowfutures.eu/pg/file/popper/view/11926/iknow-era-toolkit-2011

Resch, B., Kolokoussis, P., Hanny, D., Brovelli, M. A., & Kamel Boulos, M. N. (2025). The generative revolution: AI foundation models in geospatial health—Applications, challenges and future research. *International Journal of Health Geographics*, *24*(1), 6. https://doi.org/10.1186/s12942-025-00391-0

Saberi, M. A., Mcheick, H., & Adda, M. (2025). From Data Silos to Health Records Without Borders: A Systematic Survey on Patient-Centered Data Interoperability. *Information*, *16*(2). https://doi.org/10.3390/info16020106

Sapkota, R., Raza, S., Shoman, M., Paudel, A., & Karkee, M. (2025). Multimodal large language models for image, text, and speech data augmentation: A survey. *arXiv Preprint arXiv:2501.18648*.

Saritas, O., & Smith, J. E. (2011). The Big Picture – trends, drivers, wild cards, discontinuities and weak signals. *Special Issue: Future-Oriented Technology Analysis*, *43*(3), 292–312. https://doi.org/10.1016/j.futures.2010.11.007

Sutherland, W. J., Allison, H., Aveling, R., Bainbridge, I. P., Bennun, L., Bullock, D. J., Clements, A., Crick, H. Q. P., Gibbons, D. W., Smith, S., Rands, M. R. W., Rose, P., Scharlemann, J. P. W., & Warren, M. S. (2012). Enhancing the value of horizon scanning through collaborative review. *Oryx*, *46*(3), 368–374. https://doi.org/10.1017/S0030605311001724

Teig, C. J., Bond, M. J., Grotle, M., Kjøllesdal, M., Saga, S., Cvancarova, M. S., Ellström Engh, M. A., & Martini, A. (2023). A novel method for the translation and cross-cultural adaptation of health-related quality of life patient-reported outcome measurements. *Health and Quality of Life Outcomes*, *21*(1), 13.

UKHSA. (2025a). *Emerging infections: How and why they arise*. https://www.gov.uk/government/publications/emerging-infections-characteristics-epidemiology-and-global-distribution/emerging-infections-how-and-why-they-arise

UKHSA. (2025b). *UKHSA data dashboard*. https://ukhsa-dashboard.data.gov.uk

Urquhart, G. J., & Saunders, P. (2017). Wider horizons, wiser choices: Horizon scanning for public health protection and improvement. *Journal of Public Health*, *39*(2), 248–253. https://doi.org/10.1093/pubmed/fdw039

Villanueva-Miranda, I., Xiao, G., & Xie, Y. (2025). Artificial Intelligence in Early Warning Systems for Infectious Disease Surveillance: A Systematic Review. *Frontiers in Public Health*, *13*, 1609615.

Wei, W., Li, Y., & Si, F. (2025). Editorial: Detection and drug treatment of emerging viral diseases. *Frontiers in Cellular and Infection Microbiology*, *Volume 15-2025*. https://www.frontiersin.org/journals/cellular-and-infection-microbiology/articles/10.3389/fcimb.2025.1606637

WHO. (2021). *Ethics and Governance of Artificial Intelligence for Health: WHO Guidance* (1st ed). World Health Organization.

WHO. (2022). *Emerging trends and technologies: A horizon scan for global public health*.

WHO. (2025). *WHO Publications and Journals*. https://www.who.int/publications/journals

Willem, L., Verelst, F., Bilcke, J., Hens, N., & Beutels, P. (2017). Lessons from a decade of individual-based models for infectious disease transmission: A systematic review (2006-2015). *BMC Infectious Diseases*, *17*(1), 612. https://doi.org/10.1186/s12879-017-2699-8

Wintle, B. C., Kennicutt II, M. C., & Sutherland, W. J. (2020). Scanning horizons in research, policy and practice. In W. J. Sutherland, P. N. M. Brotherton, Z. G. Davies, N. Ockendon, N. Pettorelli, & J. A. Vickery (Eds), *Conservation Research, Policy and Practice* (pp. 29–47). Cambridge University Press; Cambridge Core. https://doi.org/10.1017/9781108638210.003

Zhao, D., Tang, Z., & He, D. (2023). A systematic literature review of weak signal identification and evolution for corporate foresight. *Kybernetes*, *53*(10), 3160–3188. https://doi.org/10.1108/K-03-2023-0343